# Moire Image Restoration using Multi Level Hyper Vision Net

D.Sabari Nathan[1], M.Parisa Beham[2] and S. M. Md Mansoor Roomi[3]
[1] Couger Inc, Tokyo, Japan.
[2]Sethu Institute of Technology,
[3]Thiagarajar college of Engineering, Madurai, India.

## Abstract

*A moire pattern in the images is resulting from high frequency patterns captured by the image sensor (colour filter array) that appear after demosaicing. These Moire patterns would appear in natural images of scenes with high frequency content. The Moire pattern can also vary intensely due to a minimal change in the camera direction/positioning. Thus the Moire pattern depreciates the quality of photographs. An important issue in demoireing pattern is that the Moireing patterns have dynamic structure with varying colors and forms. These challenges makes the demoireing more difficult than many other image restoration tasks. Inspired by these challenges in demoireing, a multilevel hyper vision net is proposed to remove the Moire pattern to improve the quality of the images. As a key aspect, in this network we involved residual channel attention block that can be used to extract and adaptively fuse hierarchical features from all the layers efficiently. The proposed algorithms has been tested with the NTIRE 2020 challenge dataset and thus achieved 36.85 and 0.98 Peak to Signal Noise Ratio (PSNR) and Structural Similarity (SSIM) Index respectively.*

## 1. Introduction

In the emerging technology era, smart phones and digital cameras provide a chance to quickly record the data which has become a vital part of our routine life. A moire pattern in the images is resulting from high frequency patterns captured by the image sensor (colour filter array) that appear after demosaicing. These Moire patterns would appear in natural images of scenes with high frequency content. These Moire patterns are the results of the intrusion between pixel grids of the camera sensor array and the digital monitor. The Moire pattern brings hostile effects on the digital images which majorly depends on the capturing angle, as well as camera shaking and reflections [1]. While image quality is continually improving, capturing superior quality natural images is still remains an issue. Moire patterns are the major reason for the contamination of digital photos as shown in Figure 1.

Thus demoireing of general dynamic images and photos is a challenging computer vision problem. Commonly, image demoire can be viewed as an image restoration task, which focus on removing noises, blurring, color disparities etc.

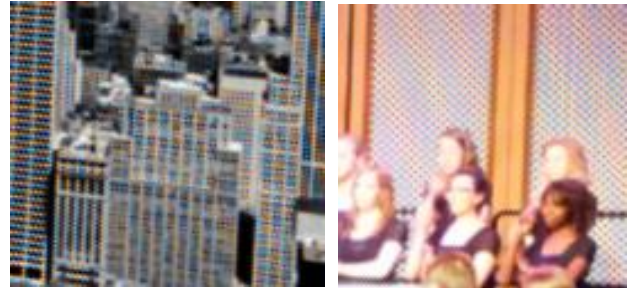

Figure 1. An example of images affected by Moire pattern

Demoireing is specifically challenging because Moire patterns are mixed with high frequency pixels in both the spatial and frequency domains. Hence, a Moire pattern could affect a high-frequency range in one image part, but a low-frequency range in another part. Thus demoireing remains challenge in any kind of digital photos irrespective of the success of various benchmark image restoration tasks, such as denoising [2] deblurring [3] and texture removal techniques [4].

Most of the conventional approaches attempt to construct blur models with simple assumptions on the sources of blurs [5]. Thus finding blur kernel is a severe problem for every pixel. Also in dynamic pictures, prediction of blur pixel is more challenging as there are several moving data. Hence, Kim et al. [6] employed a dynamic picture deblurring technique that segments and deblurs a uneven blurred image, by estimating the non-linear kernel within the segment. However, these blur kernel approximations are not well suited for rapid motion discontinuities and obstructions.

Liu et al. [7] removed Moire patterns from high-frequency kernels using a low-rank and sparse matrix decomposition model. In the literature, texture-based algorithms were also applied for the removal of Moire patterns in images having high-frequency and redundant pixels, Karacan et al. [9] considered the fact of region covariances to distinct texture features from image structure. Ono et al. [10] utilized block based texture description to decompose images into texture and structure components.

State-of-the-art methods implement variety of local filters to eradicate high-frequency textures. Nevertheless, Moire patterns in images are not only high-frequency objects but expand to a wide range of frequencies. But the texture removal algorithms fail to address the color disparities introduced by Moire patterns.

Considering the success of convolutional neural networks (CNNs), Sun et al. [8] handled the Moire patterns in both high and low frequency ranges by employing a multi-resolution CNN for automatic demoireing. Schuler et al. [11] proposed a blind deblurring method with CNN, based on conventional optimization-based deblurring methods. Sun et al. [12] proposed a sequential deblurring approach, where they generated pairs of blurry and sharp patches followed by trained the CNN to measure the possibility of a specific blur pixel of a local patch. Image restoration using most deep network models do not utilize the hierarchical features thoroughly from the moired images. Hence, Zhang et al. [20] propose a novel and efficient residual dense network (RDN) to address this problem in IR, by making a better tradeoff between efficiency and effectiveness in exploiting the hierarchical features from all the convolutional layers.

Since it was difficult for the conventional deblurring methods to approximate the blur kernels, S Nah et al. [15] proposed a multi-scale CNN that restores sharp images in a coarse to fine manner. Missing of the semantic information, degrading the performance of demoireing. To address this issue, Gao et al. [16] proposed a novel multi-scale feature enhancing network (MSFE). Furthermore, feature enhancing branch was designed to merge high-level with low-level features for forming the correlations of multiple scales. Thus, at each scale, they learned the features with richer semantic information.

Similarly, in other computer vision processes, several methods of multi-scale architecture were applied [13, 14]. However, not all multi-scale CNNs are designed to produce optimal results. Also, the major pitfalls in the existing deep learning methods are such that the feature expression ability was weak, and the characteristics of the input image couldn't be fully extracted [17]. Also transposed convolution [18] used for up sampling cannot make good use of the information and thus created artefacts in the demoired images.

Motivated by all the above issues, in this work, deep convolutional network with multilevel hyper vision net architecture is proposed to preserve fine-grained feature information. Firstly, the input image pixels are mapped to Cartesian coordinate space with the help of coordinate convolution layer. Since Moire patterns spread over high-range to low-range frequencies, our method first down samples the input image into various multilevel resolution features that contain detailed information at different levels.

Coordinate convolutional layers and the residual dense attention blocks are have also been utilized in the architecture for better performance.

As a key aspect, in this net, a multi-level supervision layers are helped to retain the multi scale hierarchical features comprised of high semantic and spatial information. These features further are then passed through a residual channel attention block which alleviate Moire patterns on the picture at various frequency bands.

Our key contributions are summarized as follows:
- An end to end attention network for enhanced demoireing is created.
- Residual Channel Attention block is introduced the in encoder and decoder.
- Convolution block attention module (CBAM)[29] is included in the skip connection which not only improve the overall mechanism, but also utilize the important features from the Encoder block.
- Multi-level supervision layers are introduced to retain the multi scale hierarchical features.
- Comprehensive experiments are conducted on the NTIRE 2020 dataset and proved the efficacy of the proposed architecture.

## 2. Proposed Hypervision Net Architecture

Inspired by the benchmark performance of deep learning methods on image restoration, a novel multilevel hyper vision net architecture is proposed that can retain the multi scale hierarchical features to eradicate Moire patterns at multiple scales in the photos or scenes. The proposed multilevel hyper vision net architecture is shown in figure 2.

In this architecture, the input image pixels are mapped to Cartesian coordinate space with the help of coordinate convolution layer. The output of coordinate convolution layers [30] are passed to encoder block. The convolutional layers and the residual dense attention blocks are utilized for better performance and to retain the multi-scale information. The proposed network has the properties of encoder and decoder structure of vanilla U-Net [18, 19]. During down-sampling three blocks have been used in the encoder phase. In each block, the first encoder block is a 3×3 convolutional layer, followed by two residual dense attention blocks are added and at the next block convolution layer with stride 2 used for down sampling. In the decoder phase, the same blocks have been used except the down sampling layer, which replaced with a super pixel [27] convolutional layer. Output features of the encoder fed to convolution block attention module (CBAM) and in the skip connection up sampled feature concatenated with encoder block, the output of the CBAM block. Inspired by Hyper Vision Net [28] model, in this work the three hyper vision layer in the decoder part is introduced.

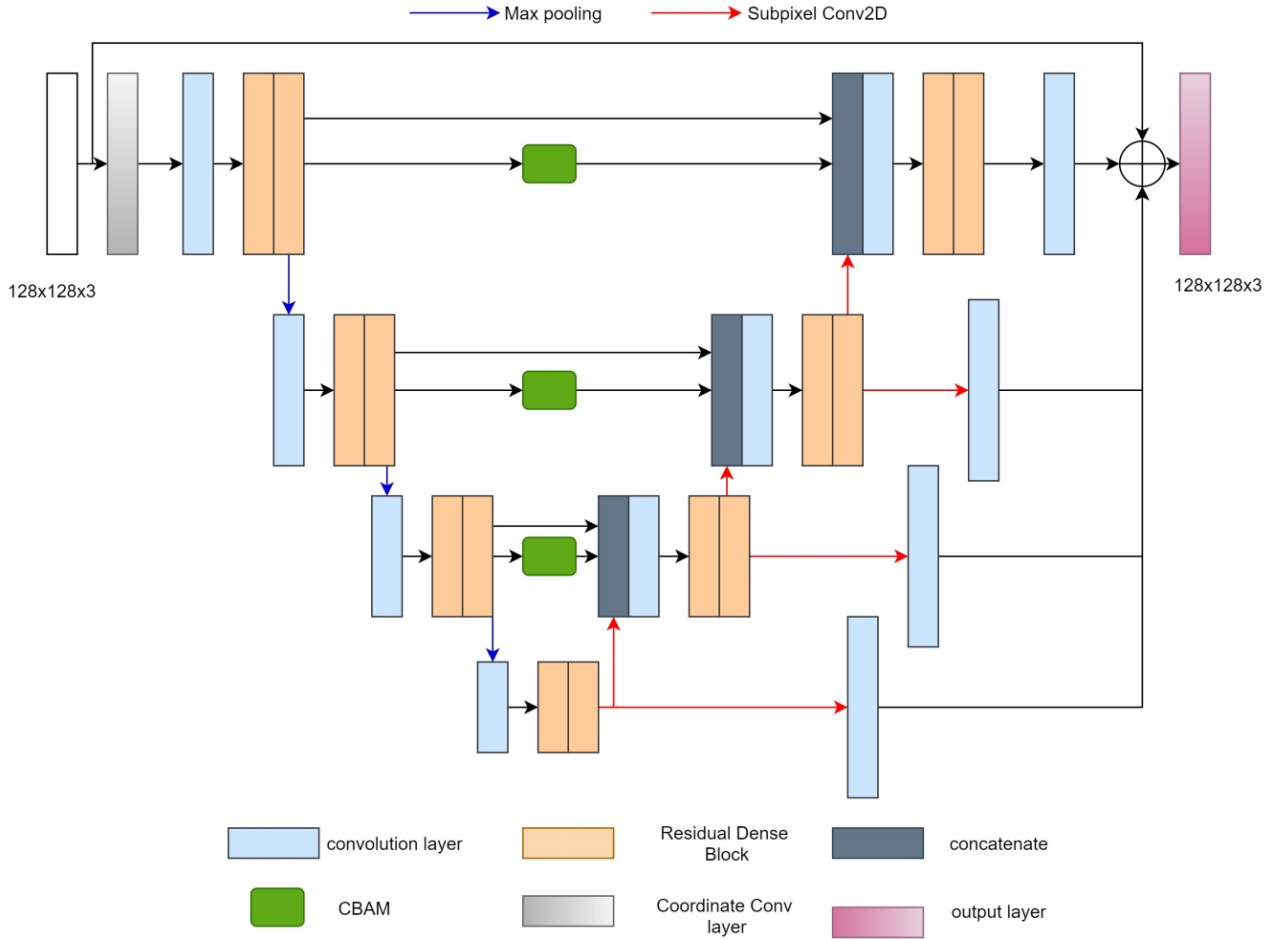

Figure 2. Proposed multilevel hyper vision net architecture

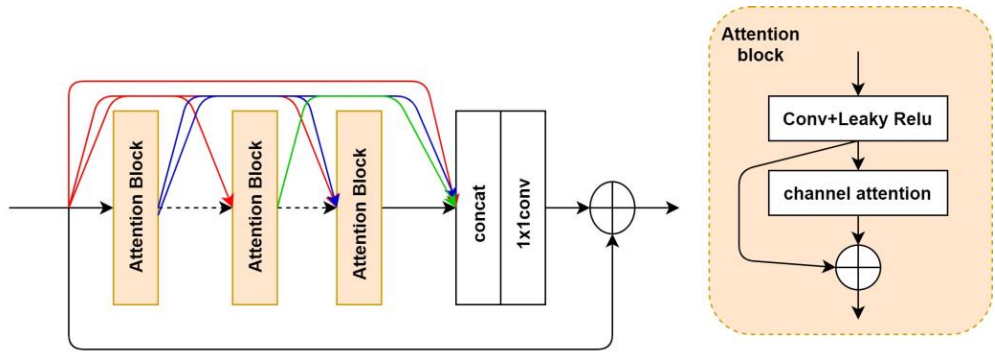

Figure 3. Proposed Residual Channel Attention Block (RCAB)

The output of these hyper vision layers are fused and supervised to obtain the enhanced demoired image. As a key aspect, in this network we involved residual channel attention block that can be used to extract and adaptively fuse hierarchical features from all the layers efficiently.

## 2.1. Residual Channel Attention Block (RCAB)

In Figure 3, our proposed Residual Channel Attention Block (RCAB) for image demoireing is illustrated. Firstly

the input image pixels are mapped to Cartesian coordinate space with the help of coordinate convolution layer.

The input image is directly given to two convolution layers to extract the low level features. These low level features are now applied to the proposed RCAB which consist of three attention blocks. The output of the three attention blocks are concatenated and convolved to obtain fine features.

As shown in Figure 3, the attention block is a composite of one convolution layer with Rectified Linear units (ReLU) [u] and a channel attention layer. Thus if the features generated from convolution layer are denoted as:

$$F_c = Conv_1(F_{cc}) \qquad (1)$$

where $Conv_1$ is the convolution operation done on the features obtained from coordinate convolution layer, $F_{cc}$. This $F_c$ act as an input to RCAB. If there are '$n$' RCAB blocks, the final output $F_n$ from the n$^{th}$ RCAB is derived as

$$F_n = P_{RCAB,n}(F_{n-1}) \qquad (2)$$

where $P_{RCAB,n}$ denotes the process of the n$^{th}$ RCAB,

$F_n$ can be observed as a local feature, since $F_n$ is produced by the n$^{th}$ RCAB fully utilizing each convolutional layers and channel attention layer within the block.

After extracting hierarchical features from the set of RCABs, global features have derived using the convolution block attention module (CBAM). Thus the global features collected from all the preceding layers

DFF makes full use of features from all the preceding layers and can be represented as:

$$F_{CB} = P_{CBAM}(F_o, F_1, F_2, ... F_{n-1}, F_{n-2}) \qquad (3)$$

where $F_{CB}$ is the output global features of CBAM by utilizing a composite functions $P_{CBAM}$.

After extracting local and global features from low level space, an up-sampling layer is loaded in the high level space. All the up sampling layers are followed by one convolution layer. Thus in this network, the residual channel attention block is used to extract and adaptively fuse hierarchical features from all the layers efficiently.

Finally the output of all the hyper vision layers in the multilevel are fused with input image and supervised to obtain the enhanced demoired image.

## 3. Experimental Results

The proposed architecture has been trained and validated with the NTIRE 2020 challenge dataset [26]. The shared NTIRE 2020 dataset consists of 10000 images. From the whole dataset 7000 images are taken for training and 3000 images are used for validation. The total Parameters of the model are 1,646,998. To prevent over fitting in our architecture, several data augmentation techniques in terms of geometric transformations are implemented. In that, the picture patches are rotated by 90 degrees, randomly flipped horizontally and vertically. Also, with the Sobel edge loss function. In addition, to supervise the model outputs and to optimize the network parameters, the model has been trained in a combination of three different loss functions such as mean squared error (MSE), SSIM Loss [25] and Sobel edge loss (SOBEL Loss) [23].

### 3.1. Optimizer and loss function

In the proposed deep network, Adam optimizer is applied which perfectly update network weights in an iterative manner in training data. Here we use learning rate from 0.001 to 0.00001 with 500 epochs.

MSE is used for maintaining the consistency between input and output and is defined as:

$$MSE = \frac{1}{N}\sum(y_{ij} - p_{ij})^2 \qquad (4)$$

$y_{ij}$ is the ground truth image and $p_{ij}$ is predicted image.

$$Loss_{SSIM} = \frac{1}{N}\sum(1 - SSIM(p_{ij})) \qquad (5)$$

where $SSIM(p_{ij})$ structural similarity index [v] for pixel p.

Sobel loss is the mean squared error of the Sobel edge information of real image and the generated image. A Sobel filter to detect the edges is applied on the generated and real image and then this information is used to calculate the mean squared error which is equal to the Sobel loss. More information can be found in [23]. A mathematical representation is given in Eqn. (6):

$$Loss_{Sobel} = \frac{1}{N}\sum_{i=1}^{N}(S(y_{ij}) - S(p_{ij}))^2 \qquad (6)$$

Thus the overall loss function is represented as:

$$LossFunction = MSE + Loss_{SSIM} + Loss_{Sobel} \qquad (7)$$

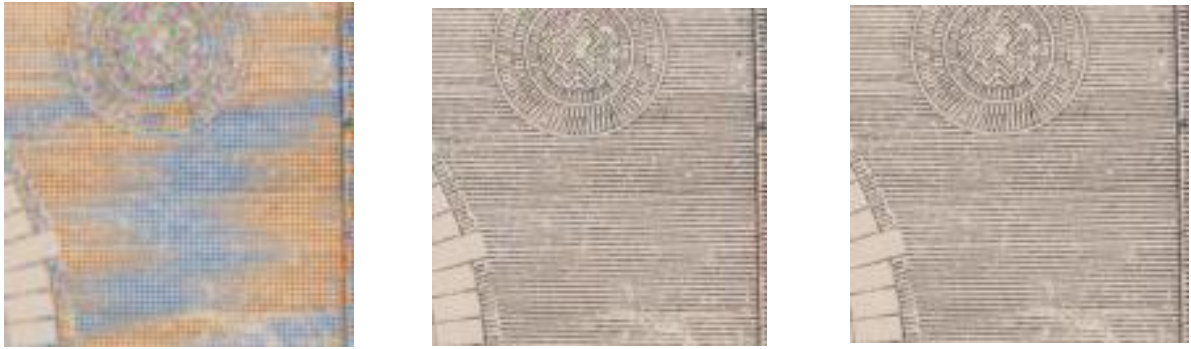

(a) (b) (c)

Figure 4. Demoired image using the proposed architecture (a) Input image (b) Ground truth image and (c) Predicted image

## 3.2. Results and Discussion

The proposed architecture is executed using Keras in Windows 10. One NVIDIA® GeForce® RTX 2070 OC with 8GB GDDR5 was used for our experimentation. Figure 4 shows the resultant predicted image after demoireing. Figure 4(a) and 4(b) shows the input and ground truth images used for training. Figure 4(c) shows the predicted image. From the figure (c) it was observed that the Moire pattern was completely removed from all the frequency range from the input image without any artefacts or noise. The qualitative result proved the efficacy of the proposed hyper vision net architecture for demoireing.

Table 1 shows the experimental results obtained using our hyper vision net architecture. For quantitative results, we considered Peak to Signal Ratio (PSNR) and Structural Similarity Index (SSIM) as a performance metrics for the evaluation of the algorithm. Peak Signal-to-Noise ratio as applied to images as a quality metric. The SSIM is a perceptual metric that quantifies image quality deprivation caused by Moire pattern. For each data the average results over all the processed images have been reported.

For 7000 training data, we achieved 36.7765 and 0.9851 PSNR and SSIM respectively. Also the proposed structure is validated with 3000 images which produced 36.2068 and 0.9831 PSNR and SSIM respectively. In the final experimentation, 500 images are tested with the proposed model and thus achieved 36.85 and 0.98 PSNR and SSIM index respectively.

Table 1. Experimental results obtained using our hyper vision net architecture

| Data size | Data | PSNR | SSIM |
|---|---|---|---|
| 7000 | Training Data | 36.7765 | 0.9851 |
| 3000 | Validation Data | 36.2068 | 0.9831 |
| 500 | Testing Data | 36.85 | 0.98 |

Table 2. Ablation study on the effects of coordinate convolution layer (CCL), CBAM and channel attention block

| Ablation effect | Train | | Validation | |
|---|---|---|---|---|
| | PSNR | SSIM | PSNR | SSIM |
| without CCL | 33.8481 | 0.97166 | 33.653 | 0.9704 |
| without CBAM | 33.9787 | 0.9719 | 33.7872 | 0.97081 |
| without channel attention | 35.3726 | 0.9798 | 35.1281 | 0.97867 |
| **Our Proposed Model** | **36.7765** | **0.9851** | **36.2068** | **0.9831** |

## 3.3. Ablation Study

Table 2 shows the ablation study on the effects of coordinate convolution layer (CCL), CBAM and channel attention. For this ablation study, first we removed the CCL from the proposed multilevel hyper vision net architecture. Now the model without CCL is trained with the same dataset with similar training and testing setup. Secondly, CBAM attention block from the skip connection is removed and the resultant architecture is trained and validated. As a final ablation effect, the channel attention block is removed from our proposed RCAB. All three effects and the resultant images are compared quantitatively and qualitatively to prove the superiority of the proposed architecture.

From the Table 3, it is observed that our proposed model compared to all the three ablation effects such as (a) Hyper vision model without the CCL, (b) Hyper vision model without CBAM and (c) Hyper vision model without Channel attention block. From the results our proposed method provides state of the art performance compared to other ablation study. Figure 5 shows the qualitative results of various ablation study effects applied on NTIRE dataset.

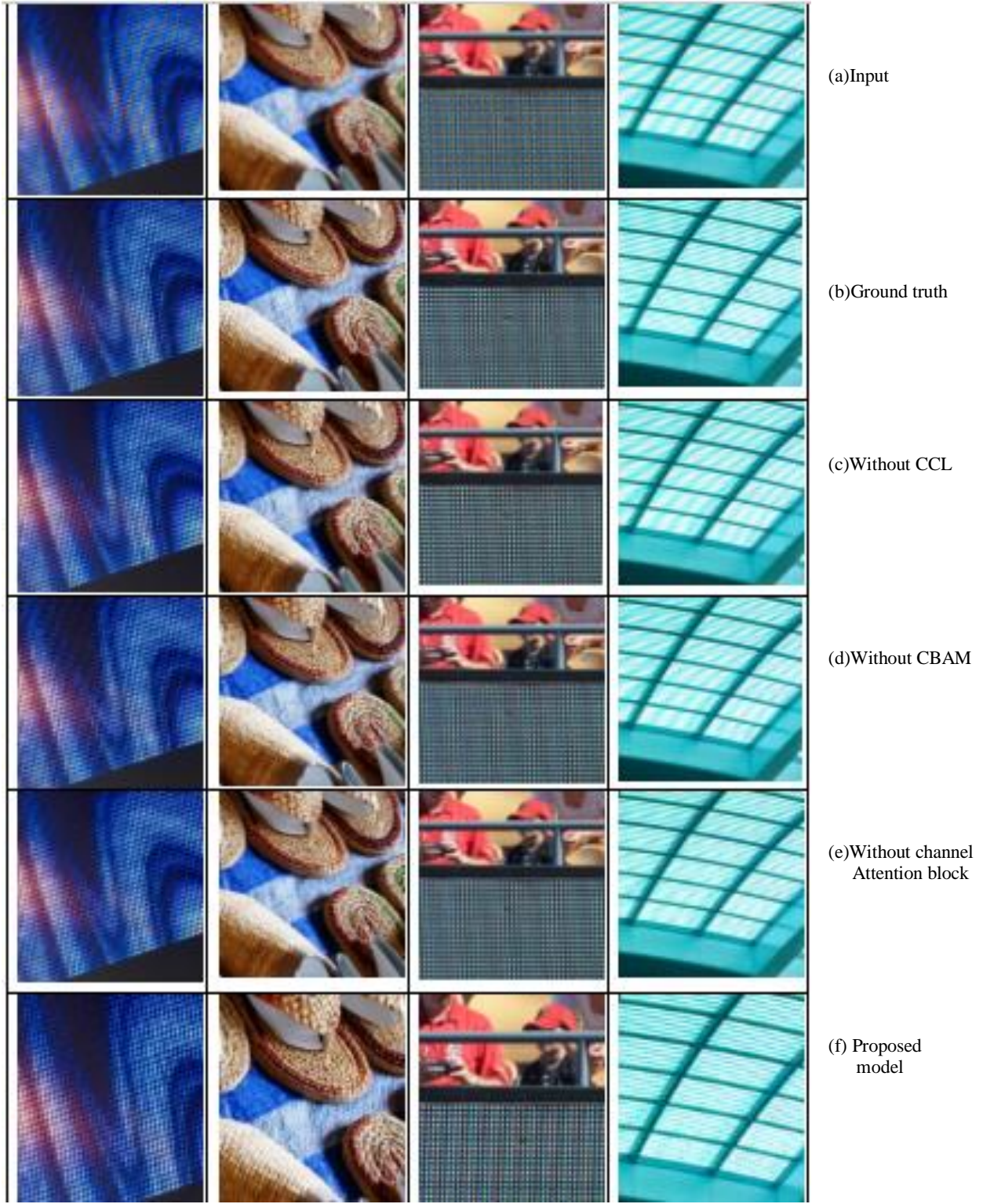

Figure 5. An example of images from NTIRE dataset shows the results of Moire pattern removal for various ablation effects. From the top to bottom rows denotes (a) Input image (b) Ground truth (c) Without CCL (d) Without CBAM (e) Without channel Attention block (f) Proposed model

Table 3. Performance comparison with standard methods with the proposed model

| Method/Metric | Sun et al. [8] | Cheng et al. [24] | MSFE [16] | Ours |
|---|---|---|---|---|
| **PSNR** | 37.41 | 42.49 | 36.66 | **36.85** |
| **SSIM** | 0.982 | 0.994 | 0.981 | **0.98** |

The resultant Moire pattern removed images are shown for various ablation effects and our proposed model. The results depicts the efficacy of our proposed model by illustrating enhanced quality by removing the Moire pattern in all the frequency range of the pictures. Table 3 shows the comparative analysis with the standard literature works with our proposed method on PSNR and SSIM. From the quantitative results it is also observed that the proposed architecture achieved good results in terms of PSNR and SSIM metrics. Thus it is proved that our hyper vision architecture is best suitable for removing the Moire pattern in all the visual images which in turn improve the quality of the images.

## 4. Conclusion

In this work, a multilevel hyper vision net deep convolutional network is proposed to remove the Moire pattern to improve the quality of the images. Here we introduced three hyper vision layers in the encoder part which are fused and supervised to obtain the enhanced demoired image in spite of varying colors. As a key contribution, in this network a residual channel attention block extracted and adaptively fused the hierarchical features from all the layers efficiently.

Compared to other literature, our model alleviates problems in digital images such as color disparities, noises and blurring. Our model was implemented in a coarse-to-fine aspect and is trained in multi-level space so as to remove the Moire pattern in high frequency as well as low frequency. The proposed algorithms has been tested with the NTIRE 2020 challenge dataset and our approach outperforms the state-of-the-art methods in both qualitative and quantitative experimentation.

## 5. References

[1]. Wikipedia. Moir pattern. https://en.wikipedia.org/wiki/Moir%C3%A9_pattern.
[2]. Tianyu Gao, Yanqing Guo, Xin Zheng, Qianyu Wang, and Xiao-Jiao Mao, Chunhua Shen, and Yu-Bin Yang, "Image restoration using very deep convolutional encoder decoder networks with symmetric skip connections," in Advances in Neural Information Processing Systems 29:Annual Conference on Neural Information Processing Systems 2016, December 5-10, 2016, Barcelona, Spain, 2016, pp. 2802–2810.
[3]. Seungjun Nah, Tae Hyun Kim, and Kyoung Mu Lee, "Deep multi-scale convolutional neural network for dynamic scene deblurring," in 2017 IEEE Conference on Computer Vision and Pattern Recognition, CVPR 2017, Honolulu, HI, USA, July 21-26, 2017, 2017, pp. 257–265.
[4]. H. Cho, H. Lee, H. Kang, and S. Lee, "Bilateral texture filtering," ACM Transactions on Graphics (TOG), vol. 33, no. 4, p. 128, 2014.
[5]. Gupta, N. Joshi, C. L. Zitnick, M. Cohen, and B. Curless. Single image deblurring using motion density functions. In ECCV, pages 171–184. Springer, 2010.
[6]. T. H. Kim, B. Ahn, and K. M. Lee. Dynamic scene deblurring. In ICCV, 2013.
[7]. Fanglei Liu, Jing-Yu Yang, and Huanjing Yue, "Moiré pattern removal from texture images via low-rank and sparse matrix decomposition," in 2015 Visual Communications and Image Processing, VCIP 2015, Singapore, December 13-16, 2015, 2015, pp. 1–4.
[8]. Yujing Sun, Yizhou Yu, and Wenping Wang, "Moiré photo restoration using multi-resolution convolutional neural networks," IEEE Trans. Image Processing, vol. 27, no. 8, pp. 4160–4172, 2018.
[9]. L. Karacan, E. Erdem, and A. Erdem, "Structure - preserving image smoothing via region covariances," ACM Transactions on Graphics (TOG), vol. 32, no. 6, p. 176, 2013.
[10]. S. Ono, T. Miyata, and I. Yamada, "Cartoon-texture image decomposition using blockwise low-rank texture characterization," IEEE Transactions on Image Processing, vol. 23, no. 3, pp. 1128–1142, 2014.
[11]. J. Schuler, M. Hirsch, S. Harmeling, and B. Schölkopf. Learning to deblur. IEEE transactions on pattern analysis and machine intelligence, 8(7):1439–1451, 2016.
[12]. J. Sun, W. Cao, Z. Xu, and J. Ponce. Learning a convolutional neural network for non-uniform motion blur removal. In CVPR, pages 769–777. IEEE, 2015.
[13]. Eigen, C. Puhrsch, and R. Fergus. Depth map prediction from a single image using a multi-scale deep network. In Advances in Neural Information Processing Sytems, pages 2366–2374, 2014.
[14]. Eigen and R. Fergus. Predicting depth, surface normal and semantic labels with a common multi-scale convolutional architecture. In ICCV, pages 2650–2658, 2015.
[15]. Seungjun Nah, Tae Hyun Kim, Kyoung Mu. Lee. Deep Multi-scale Convolutional Neural Network for Dynamic Scene Deblurring. 2017 IEEE Conference on Computer Vision and Pattern Recognition, 2017.
[16]. Tianyu Gao, Yanqing Guo, Xin Zheng, Qianyu Wang, and Xiangyang Luo. Moiré pattern removal with multi-scale feature enhancing network. In 2019 IEEE International Conference on Multimedia & Expo Workshops (ICMEW), pages 240–245. IEEE, 2019.
[17]. Zhenli Zhang, Xiangyu Zhang, Peng Chao, Xiangyang Xue, and Sun Jian. Exfuse: Enhancing feature fusion for semantic segmentation. 2018.
[18]. Matthew D Zeiler, Dilip Krishnan, Graham W Taylor, and Rob Fergus. Deconvolutional networks. In 2010 IEEE Computer Society Conference on computer vision and pattern recognition, pages 2528–2535. IEEE, 2010.

[19]. Bolin Liu, Xiao Shu, and Xiaolin Wu. Demoireing of camera-captured screen images using deep convolutional neural network. arXiv preprint arXiv:1804.03809, 2018.
[20]. Yulun Zhang, Yapeng Tian, Yu Kong, Bineng Zhong, and Yun Fu, "Residual Dense Network for Image Restoration", arXiv: 1812.10477 [cs.CV].
[21]. X. Glorot, A. Bordes, and Y. Bengio, "Deep sparse rectifier neural networks," in Proc. International Conference on Artificial Intelligence and Statistics, 2011.
[22]. Zhou Wang, Alan C Bovik, Hamid R Sheikh, and Eero P Simoncelli. Image quality assessment: from error visibility to structural similarity. IEEE transactions on image processing, 13(4):600–612, 2004.
[23]. Shanxin Yuan, Radu Timofte, Gregory Slabaugh, Ales Leonardis, Bolun Zheng, Xin Ye, Xiang Tian, Yaowu Chen, Xi Cheng, Zhenyong Fu, et al. Aim 2019 challenge on image demoireing: Methods and results. arXiv preprint arXiv:1911.03461, 2019.
[24]. Xi Cheng, Zhenyong Fu and Jian Yang. Multi-scale Dynamic Feature Encoding Network for Image Demoir´eing. arXiv: 1909.11947v1 [cs.CV] 26 Sep 2019.
[25]. Bolin Liu, Xiao Shu, and Xiaolin Wu. Demoirn'eing of camera - captured screen images using deep convolutional neural network. arXiv preprint arXiv:1804.03809, 2018.
[26]. Shanxin Yuan, Gregory Slabaugh, and Radu Timofte. NTIRE 2020 Demoireing Challenge Factsheet.2020.
[27]. W. Shi *et al*., "Real-Time Single Image and Video Super-Resolution Using an Efficient Sub-Pixel Convolutional Neural Network," *2016 IEEE Conference on Computer Vision and Pattern Recognition (CVPR)*, Las Vegas, NV, 2016, pp. 1874-1883.
[28]. D.Sabarinathan, M.Parisa Beham, S.M.Md.Mansoor Roomi, "Hyper Vision Net: Kidney Tumor Segmentation Using Coordinate Convolutional Layer and Attention Unit", Image and Video Processing (eess.IV).2019,arXiv:1908.03339
[29]. Woo, Sanghyun, et al. "CBAM: Convolutional block attention module." Proceedings of the European Conference on Computer Vision (ECCV). 2018.
[30]. Liu R, Lehman J, Molino P, Such FP, Frank E, Sergeev A, Yosinski J. An intriguing failing of convolutional neural networks and the coordconv solution. In Advances in Neural Information Processing Systems, pp. 9605-9616, 2018.